\documentclass[journal]{IEEEtran}
\usepackage{cite}

\ifCLASSINFOpdf
  \usepackage[pdftex]{graphicx,graphics}
  \DeclareGraphicsExtensions{.pdf,.jpeg,.png}
\else
  \usepackage[dvips]{graphicx,graphics}
  \DeclareGraphicsExtensions{.eps}
\fi

\graphicspath{{figures/}}

\usepackage[cmex10]{amsmath}
\usepackage{amssymb}
\usepackage{algorithmic}
\usepackage{algorithm}

\usepackage{array}
\usepackage{multirow}

\usepackage[caption=false,font=footnotesize]{subfig}
\usepackage{float}

\usepackage{url}

\usepackage{tikz}

\usepackage{xcolor}
\usepackage{multirow}
\usepackage{color, colortbl}

\usepackage{listings}

\definecolor{Gray}{gray}{0.9}
\definecolor{LightGray}{gray}{0.8}
\definecolor{LightCyan}{rgb}{0.88,1,1}
\definecolor{LawnGreen}{rgb}{0.48,0.98,0}
\definecolor{mygreen}{RGB}{28,172,0} 
\definecolor{mylilas}{RGB}{170,55,241}

\usepackage{lipsum}

\begin{document}

\title{Safe, Remote-Access Swarm Robotics Research on the Robotarium}
%
%
%

\author{Daniel~Pickem,
	Li~Wang,
	Paul~Glotfelter,
	Yancy~Diaz-Mercado,
	Mark~Mote,
	Aaron~Ames,
	Eric~Feron, 
	and~Magnus~Egerstedt

\thanks{This research was sponsored by Grants No. 1531195 and 1544332 from the U.S. National Science Foundation.} 
\thanks{The authors are with the Georgia Institute of Technology, Atlanta, GA 30332, USA, \{daniel.pickem,liwang,paul.glotfelter,yancy.diaz,mmote3,ames,\newline  feron,magnus\}@gatech.edu.}%
}


\markboth{Manuscript Draft, April~2016}%
{Shell \MakeLowercase{\textit{et al.}}: Bare Demo of IEEEtran.cls for Journals}


\maketitle

\begin{abstract}
This paper describes the development of the \textit{Robotarium} -- a remotely accessible, multi-robot research facility. The impetus behind the Robotarium is that multi-robot testbeds constitute an integral and essential part of the multi-agent research cycle, yet they are expensive, complex, and time-consuming to develop, operate, and maintain. These resource constraints, in turn, limit access for large groups of researchers and students, which is what the Robotarium is remedying by providing users with remote access to a state-of-the-art multi-robot test facility. This paper details the design and operation of the Robotarium as well as connects these to the particular considerations one must take when making complex hardware remotely accessible. In particular, safety must be built in already at the design phase without overly constraining which coordinated control programs the users can upload and execute, which calls for minimally invasive safety routines with provable performance guarantees.
\end{abstract}


%
\IEEEpeerreviewmaketitle

\section{Introduction}
\label{sec:introduction}
Coordinated control of multi-robot systems has received significant attention during the last decade, with a number of distributed control and decision algorithms being developed to solve a wide variety of tasks, ranging from environmental monitoring (e.g., \cite{Cortes2004,Dantu2005,Zhu2013}) to collective material handling (e.g., \cite{Petersen2011}). These developments have been driven by a confluence of 
algorithmic advances, increased hardware miniaturization, and cost reduction, and a number of compelling multi-robot testbeds have been developed. However, despite these advances, it is still a complex and time-consuming proposition to go from theory and simulation, via a small team of robots, all the way to a robustly deployed, large-scale multi-robot system, and there are only a handful of laboratories around the world that can field massive numbers of robots in the air, under water, or on the ground, e.g., \cite{Johnson2006, Michael2008a, De2006, Rubenstein2012, Rubenstein2013}.

Despite the relative sparsity of large-scale facilities, contributions to the field have been made in fields as disparate as networked control theory, robotics, biology, game theory, and even sociology, e.g., \cite{pinciroli2011,Couzin2005,wilson2014,zhu2015,stojmenovic2014,hills2015}, where simulated robots serve as a proxy for their physical counterparts.  However, to advance multi-robot research further, actual deployment is crucial since it is increasingly difficult to faithfully simulate all the issues associated with making multiple robots perform coordinated tasks. This is due to the increased task complexity, perceptual occlusions and cross-talk effects, and saturation of communication channels that follow from increased robot density. 

In response to this theory-simulation-practice gap, the Robotarium is an open, remote-access multi-robot system, explicitly designed to address these issues by providing access that is flexible enough to allow for a number of different scientific questions to be asked, and different coordination algorithms to be tested. And, although a number of elegant, remote-access robot systems have been developed in the past (e.g., \cite{Johnson2006, De2006, Pitzer2012, Casan2015}), what makes the Robotarium different is its explicit focus on supporting \textit{multi-robot research}, as opposed to, for example, educational or single-robot systems. 

At its core, the Robotarium is a multi-robot laboratory, where mobile robots can coordinate their behaviors in a collaborative manner with guaranteed safety, i.e., the physical assets of the Robotarium are protected by guaranteed collision avoidance in a minimally invasive manner. In this paper, we discuss how this multi-robot laboratory is structured and, in particular, how the explicit focus on being a safe remote-access research platform informs the design. In fact, the report from a recent NSF Workshop on Remotely Accessible Testbeds \cite{Egerstedt2015} identified this inherent safety/flexibility tension as one of the key questions when pursuing a "science of remote access".

%

The outline of the paper is as follows. Section \ref{sec:relatedWork} presents an overview of currently available (remote-access) testbeds in the (multi-)robot and sensor networks domain. Section \ref{sec:robotarium} introduces the Robotarium design concept, while Section \ref{sec:safety} addresses the challenge of safety, or more precisely how collisions can be avoided in a minimally invasive fashion while affording users the maximum flexibility in executing their custom code. Section \ref{sec:experiments} presents a number of swarm-robotic experiments instantiated on the Robotarium to highlight the breadth of tasks that can be accomplished. 
Finally, Section \ref{sec:conclusion} concludes the paper and elaborates on future work.

\begin{figure*}[tbp]
  \begin{center}
    \includegraphics[width=1.0\textwidth]{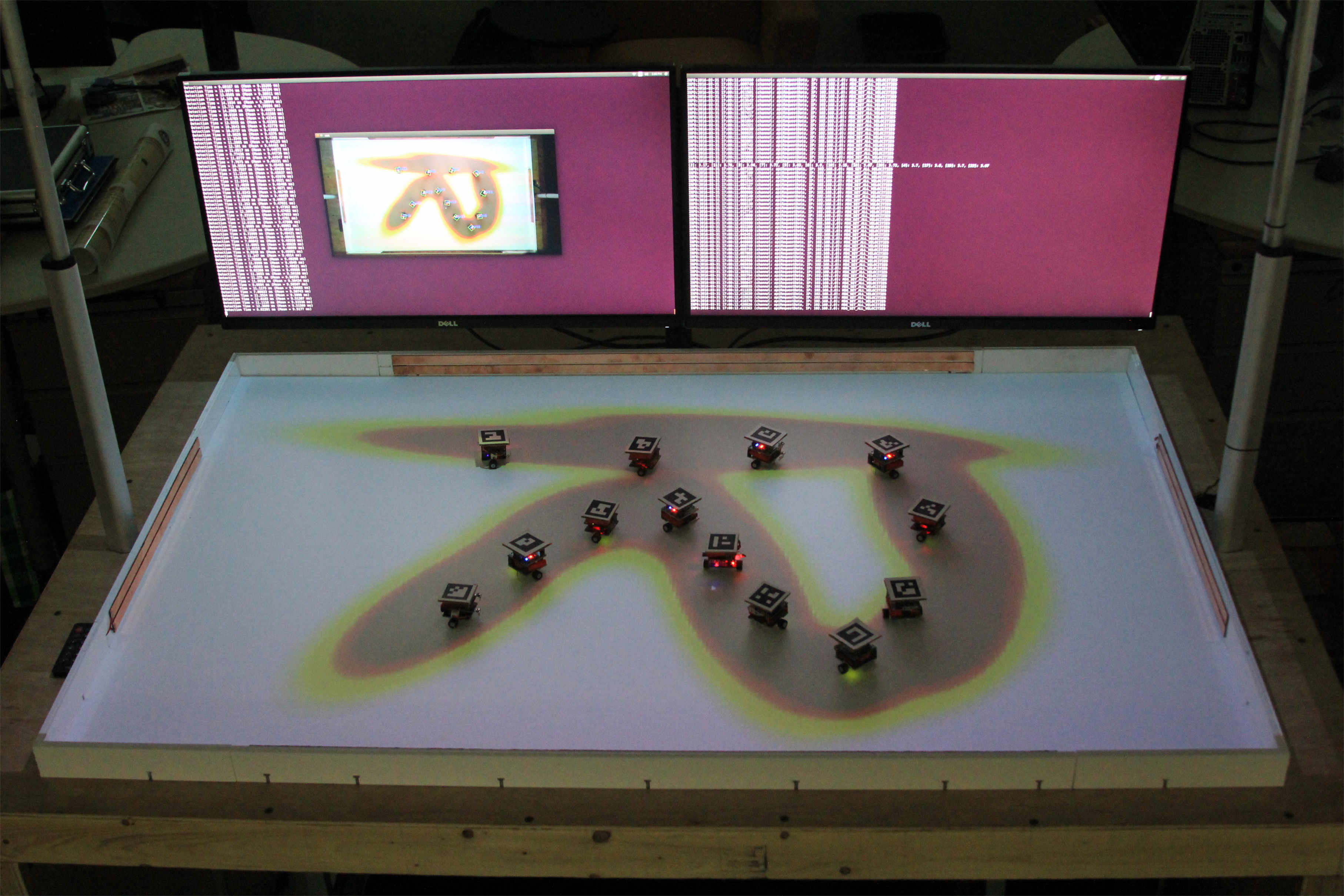}
    \caption{Example of a coverage control algorithm executed on the Robotarium using 13 GRITSBot robots. The desired density function is projected onto the testbed arena in the shape of an R.}
    \label{fig:static_coverage_control_title_page}
  \end{center}
\end{figure*}


\section{Related Work}
\label{sec:relatedWork}
The Robotarium's main goal is to provide researchers and students access to a remotely-accessible multi-robot testbed. In this section, we briefly survey the field of remote access testbeds of relevance to the Robotarium vision, and broadly categorize them along the following dimensions: multi-agent robotic testbeds, sensor network testbeds, and remotely accessible educational tools. A comprehensive overview of such testbeds can also be found in \cite{Groeber2007,Jimenez-Gonzalez2013}.
  
\subsection{Robotic Testbeds}
In the world of shared robotics testbeds, only a handful are publicly available. One example of a shared robotic platform is the PR2 Remote Lab at Brown University (e.g., \cite{Osentoski2011,Osentoski2012,Pitzer2012}), which makes a single PR2 robot available for public use. Similar humanoid robots have also been available for the purpose of crowdsourced learning from demonstration in \cite{Penaloza2013,Toris2014a,Toris2015}. It is worth noting that the safe execution of experiments in these research efforts depended on compliant actuators or the presence of an experimenter in the lab to halt the robot in case of impending collisions.

Remotely accessible multi-robot testbeds designed for research purposes were proposed in \cite{Michael2008a, Michael2011} and also in \cite{Cruz2007}. The latter testbed, called COMET, was developed at Oklahoma State University and relied on a small numbers of robots. While offering similar capabilities to the Robotarium, the robots provided by these testbeds occupy a significantly larger footprint and are considerable more expensive than the Robotarium robots, which is an inherent obstruction to reaching large numbers of robots.
Similar mobile robot testbeds have also emerged from the wireless sensor network domain. The Mobile Emulab \cite{Johnson2006}, for example, closely resembles the Robotarium in that it provides a shared, remotely accessible multi-agent testbed with mobile nodes. However, its focus lies on sensor networks and evaluating mobility-related network protocols. Thus, the Mobile Emulab uses mobility only as a means to establish static sensor network topologies and not as an inherent component of dynamic multi-agent experiments.

Finally, originating in the educational domain, Robotnacka is a robotic laboratory offering remote access to three robots \cite{Petrovic2012}. It is an educationally focused resource, available 24/7 through an online interface, has automatic charging through docking stations, and allows virtual robots to be used. Extending this system to a whole swarm of robots, however, would require significant amounts of space due to the robots' large footprint. Most importantly, however, Robotnacka relies on well-intentioned users as it supports no safety measures to avoid inter-robot collisions.
  
A number of effective, yet non-remotely-accessible testbeds have been developed for a number of locomotion modalities. For example, hovercraft vehicles have been used in testbeds at Caltech \cite{Jin2004} and the University of Illinois \cite{Vladimerou2006}. Testbeds at the University of Pennsylvania rely on micro UAVs \cite{Michael2010} as well as a combination of ground and aerial vehicles \cite{Chaimowicz2004, Grocholsky2006}. Similarly, the multi-vehicle testbed at MIT \cite{King2004} and the testbed at Brigham Young University \cite{Mclain2004} use a combination of ground and aerial vehicles. Although these testbeds are in principle capable of remote operation, the required infrastructure is not available. 
Among the variety of inexpensive robotic testbeds that have been presented in the literature one stands out above all for the sheer scale -- the Kilobot testbed. Although it is a closed, non-shared testbed, it is used for large scale experiments in 2D self-assembly and collective transport with up to 1024 robots (e.g., \cite{Rubenstein2012, Rubenstein2013}). However, the Kilobot's locomotion modality (vibration motors) makes it less suitable for a general purpose multi-robot testbed focusing on coordinated control as opposed to distributed computation. 
  
Unlike these testbeds, which rely on expert knowledge to operate and protect the physical assets from damage, the Robotarium aims to be simple to use and interface with, yet at the same time be inherently safe to operate. In other words, built-in safety measures prevent users from causing accidental or purposeful damage to the robots. 
  
\subsection{Sensor Networks Testbeds and Cyber Security}
Some of the earliest remote-access testbeds resided in the sensor networks and cyber security domains. Limiting access to largely immobile computing and sensing nodes mitigated the risk of making them publicly accessible. This category includes testbeds such as PlanetLab \cite{Chun2003}, MoteLab \cite{Werner-Allen2005}, MiNT \cite{De2005} and MiNT-m \cite{De2006}, ORBIT \cite{Raychaudhuri2005}, and more recently DeterLab \cite{Mirkovic2012} and CONET \cite{Martinez-deDios2013}. 
Two testbeds that stand out for their long period of activity are the ORBIT testbed \cite{Raychaudhuri2005} and the DeterLab \cite{Bajcsy2004, Mirkovic2012}. These instruments have enabled research in the sensor networks and cyber security domain for over ten years. However, the nodes in these systems are stationary and as such fulfill a distinctly different purpose than the mobile robots of the Robotarium. 
Unlike most other sensor network testbeds, only MiNT-m \cite{De2006} and CONET \cite{Martinez-deDios2013} contain mobile sensing nodes. While MiNT-m's tethered mobile nodes enable just enough mobility to automatically establish static sensing topologies, CONET combines wireless sensor networks with mobile robots in a more general fashion. However, the resource requirements of the latter testbed are significant: a total area of 500m$^2$ housing six large mobile robots and numerous stationary sensor nodes. 

\subsection{Educational Testbeds}
A number of testbeds have originated in the educational domain with the purpose of facilitating and enriching the learning experience in areas such as physics \cite{Groeber2007}, embedded systems and microcontroller programming \cite{Seiler2012, Sell2012, Sell2012a,Sell2013}, robotic manipulation \cite{Marin2003,Lopez2007,Latinovic2015}, mobile robot coordination \cite{Sell2008,Kodagoda2013}, and even humanoid robot programming \cite{Casan2015}.
For example, the \textit{Robotic Programming Network (RPN)} \cite{Casan2015} has recently made a single Nao robot available for remote experimentation. Similarly, a remote robotics lab discussed in \cite{Kodagoda2013} makes a single robot available for localization and path planning experiments, but still relies heavily on simulation. While providing the required infrastructure for remote access, compared to the Robotarium, two main drawbacks exist: only a single robot is available in each case and neither approach mentions the critical safety aspect.
While some work especially in the educational domain has already hinted at safety and security issues \cite{Wirz2006, Wirz2009}, code verification and validation techniques as well as collision avoidance guarantees are still lacking in most testbeds. 

What sets the Robotarium apart from these efforts in the educational domain is it's explicit focus on being a state-of-the-art \textit{research} instrument that allows the flexible execution of custom user code, while at the same time guaranteeing the safety of the Robotarium's physical assets. 
Another aspect of remote access that the Robotarium tackles in particular is its always on-nature that allows robots to operate 24/7 through autonomous recharging and automated system management. This continuous operation emphasizes the need for provable collision avoidance even more since even small probabilities of collision will over time lead to collisions with near-certainty.

\section{The Robotarium}
\label{sec:robotarium}
The Robotarium, unlike the testbeds discussed in the previous section, incorporates aspects of remote operation that are unique in the way they are combined. What the Robotarium ultimately aims to be is a large-scale swarm-robotic testbed that is accessible around the clock, gives users the flexibility to test any algorithm they wish, and evolves in response to changing user needs. In particular the Robotarium tackles the challenge of robust and safe operation of a large collection of robots with minimal operator intervention and maintenance. The always on-nature of the Robotarium highlights the need for automated maintenance, which relies on robust position tracking, automated battery recharging, and collision-free execution of motion paths. In this section, we will outline how these requirements guide the design of the first prototype and elaborate on both the hardware as well as software architecture of the Robotarium.
  

\subsection{Design Considerations}
\label{subsec:design_considerations}
As a shared, remotely accessible, multi-robot facility, the Robotarium's main purpose is to lower the barrier of entrance into multi-agent robotics. Similar to open-source software that provides access to high quality software, the Robotarium 
will have to exhibit a subset or all of the following high-level characteristics to fulfill its intended use effectively.

\begin{itemize}
    \item Enable students and researchers to simply and inexpensively replicate the Robotarium testbed and its robots, implying that the design needs to be simple and open-source.
    \item Enable intuitive interaction with and simple data collection from the Robotarium.
    \item Enable a seamless switch between developing algorithms in simulation and execution of the same algorithms on the physical robots.
    \item Minimize the cost and complexity of maintaining a large collective of robots.
    \item Keep the system extensible in terms of adding more robots as well as different types of robots.
    \item Integrate safety and security measures to protect the Robotarium from damage and misuse.
\end{itemize}

These desired high-level characteristics can be mapped onto more specific constraints that inform the hardware design as well as the software architecture. The current instantiation of the Robotatium satisfies these requirements as follows:

\begin{itemize}
    \item Large numbers of low-cost robots (currently up to 20 robots are made available)
    \item Convenience features to simplify the maintenance of large collectives of robots (automatic charging and tracking)
    \item Public interface enabling simple code submission and data/video retrieval from the testbed
    \item Built-in safety features that guarantee collision avoidance (see Section \ref{sec:safety})
\end{itemize}

\subsection{The Science of Remote Access}
\label{subsec:science_of_remote_access}
Beyond the logistics of providing access, scheduling and managing remote users, and providing meaningful experimental data, the development of remotely accessible research platforms comes with a number of challenges that their educational counterparts do not share. In the Report from the US National Science Foundation Workshop on Accessible, Remote Testbeds \cite{Egerstedt2015}, the "science of remote access" was discussed. One key problem that must be resolved concerns the inherent trade-off between flexibility and safety.

For the Robotarium to be a truly useful research instrument, users must be able to push the boundaries for what the robots can achieve. In fact, the Robotarium must be designed to support algorithms and inquiries that have yet to be imagined, implying that the allowable control programs cannot be needlessly constrained. But, at the same time, users should not be allowed to break the robots through truly reckless maneuvers. As such, collisions must be avoided in a minimally invasive manner, allowing for the user-provided commands to control the system as much as possible, subject to safety constraints.

In this paper, we describe how this balance between flexibility and safety is struck using control barrier certificates. By defining the set of safe states, barrier certificates are constructed that, as long as they are satisfied, the robots are guaranteed to be safe. The actual control signals then minimizes the distance to the user specified control signals subject to the constraint that the certificates are satisfied, and the resulting system is safe in a provably minimally invasive manner.

An additional complication is that users may want to test different aspects multi-agent robotics, such as verifying behavioral models of social insects or testing formation controllers for nonholonomic multi-agent systems. What this implies is that the Robotarium must be accessible at different levels of abstraction, in terms of the robot dynamics, the sensing capabilities, and even the information exchange network supported by the system. This is another key aspect of the "science of remote access" that will be discussed in this paper.

\begin{figure*}[tbp]
  \begin{center}
    \includegraphics[width=1\textwidth]{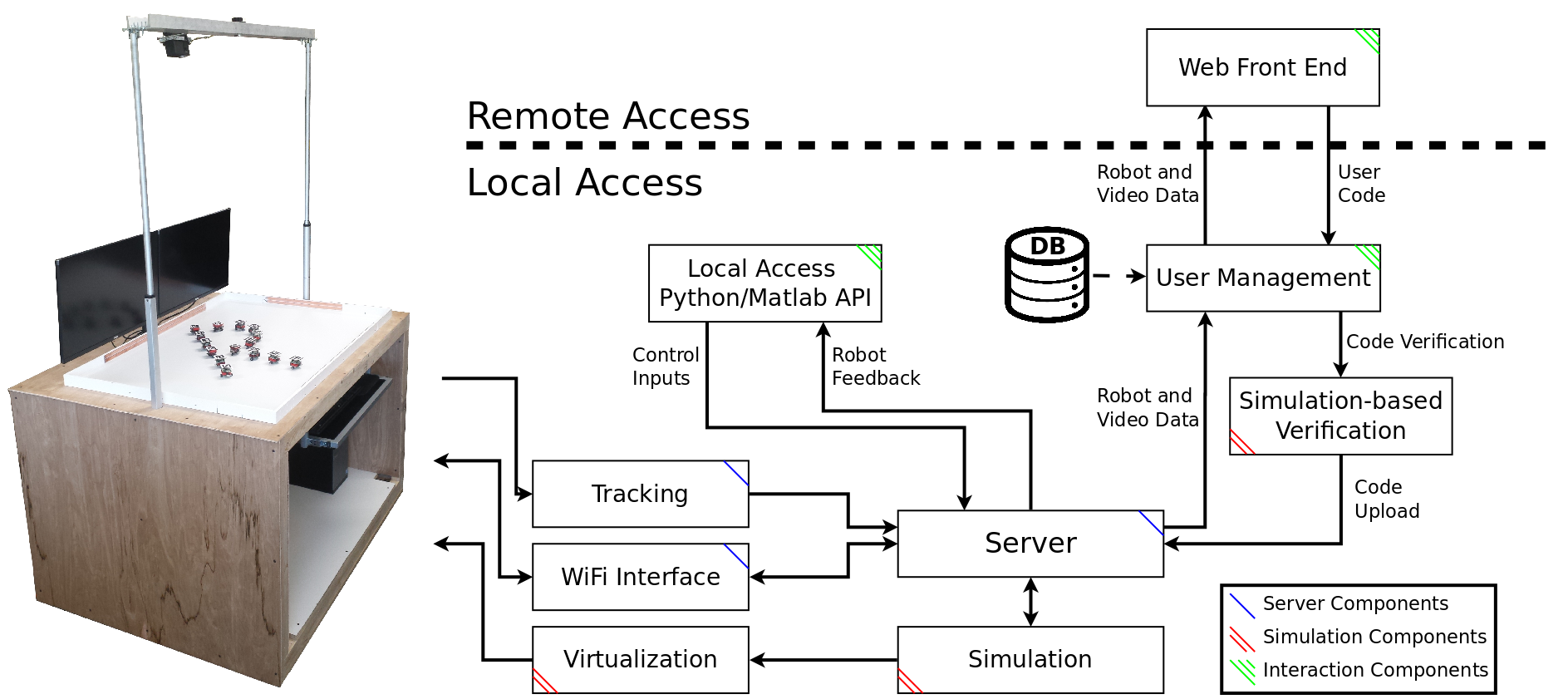}
    \caption{System architecture overview. The Robotarium prototype includes components that are executed locally on Robotarium infrastructure as well as user-facing components that run on remote user machines (web front end). Three components interact directly with the robot hardware -- tracking, wireless communication, and virtualization. The remaining components handle user management, code verification and upload to the server, as well as coordination of user data and testbed-generated data.}
    \label{fig:system_architecture}
  \end{center}
\end{figure*}

\subsection{Prototype Design}
\label{subsec:prototype_design}
In this section, we will elaborate on the hardware and software components of the Robotarium as well as their interaction with each other, the robots, and the users (see Fig. \ref{fig:system_architecture}).\footnote{All design and code files are open-source and can be found at \url{www.robotarium.org}.} The Robotarium hardware includes the robots themselves, the position tracking system based on web cameras, wireless communication hardware, as well as a charging system built into the walls of the arena. The software stack consists of the coordinating server application, APIs, simulation and virtualization infrastructure, as well as a simulation-based code verification block. Note that the shown architecture represents the current development snapshot of the Robotarium. By design, however, the Robotarium has to evolve over time in response to user needs in order to provide an effective research instrument as opposed to a static showcase.

\subsubsection{Hardware}
\label{subsubsec:hardware}
The Robotarium is meant to provide a well integrated, immersive user experience with the smallest possible footprint, and features that allow a large swarm to be maintained effortlessly. Such tight integration is only possible with custom hardware. At the core of the Robotarium are therefore our custom-designed robots - the GRITSBots (introduced in \cite{Pickem2015}). 
The tightly integrated design of the GRITSBot and the Robotarium allows a single user to easily operate and maintain a swarm of robots through features such as (i) automated registration with the overhead tracking system, (ii) automatic battery charging, (iii) wireless (re)programming, and (iv) automatic sensor calibration. These features have been described in detail in \cite{Pickem2015}. However, given their importance for the operation of a remote access testbed, we review the aspects of global position tracking and automatic recharging in detail.
    
\begin{itemize}
\item{\textit{Robots:}}
The Robotarium leverages the GRITSBot, a miniature wheeled ground robot that we introduced in \cite{Pickem2015}. The GRITSBot is an inexpensive differential drive miniature robot featuring a modular design that permits its hardware capabilities to be adapted easily to different tasks. A key feature that makes the GRITSBot the basis for the Robotarium is that it enables a straightforward transition from typical experimental multi-robot setups to GRITSBot-based experiments because the GRITSBot closely resembles popular platforms in capabilities and architecture. In particular, the robot's main features include (i) high accuracy locomotion through miniature stepper motors, (ii) range and bearing measurements through infrared distance sensing, (iii) global positioning system through an overhead camera system, and (iv) WiFi-based communication with a global host. The detailed specifications can be found in \cite{Pickem2015} while design updates and design files are available at \url{www.robotarium.org}.
      

\begin{figure}[tbp]
	\centering
	\subfloat[Robot showing all three layers (from top to bottom): sensor, main, and motor layer.]{\includegraphics[width=0.2\textwidth]{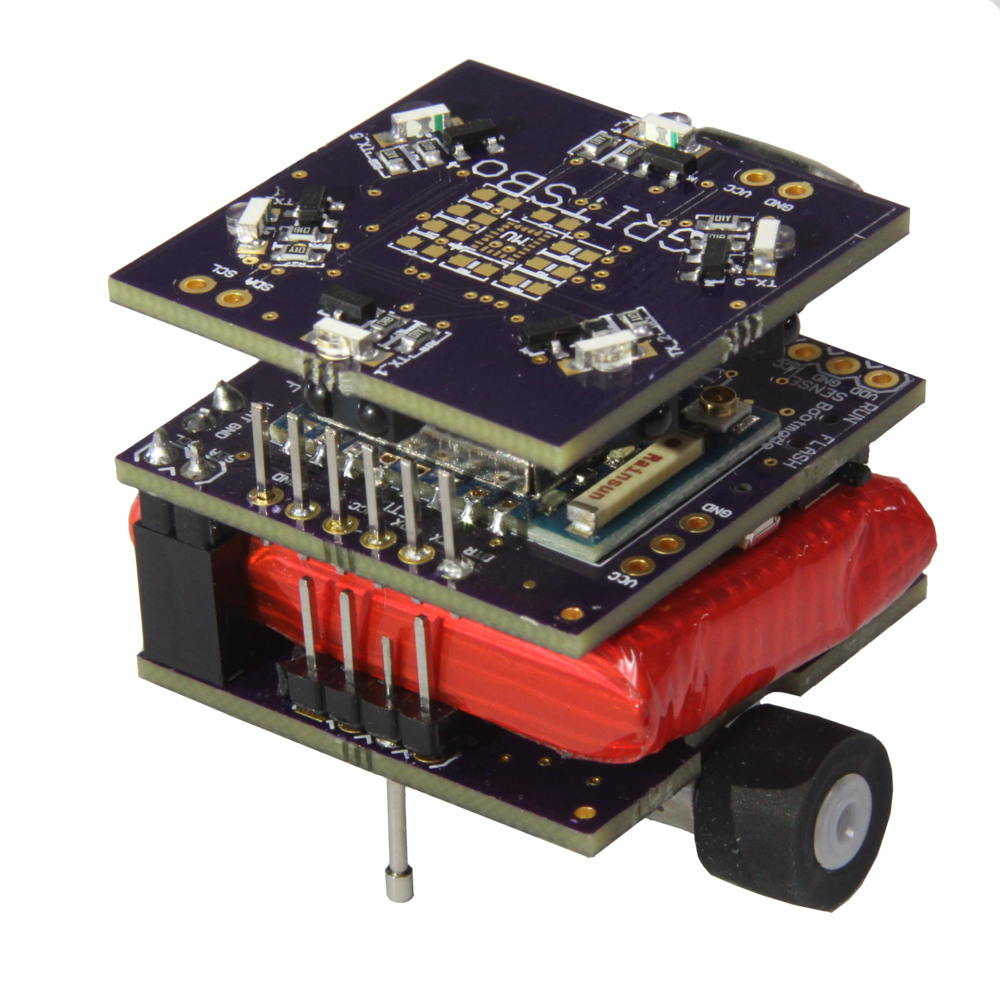} \label{fig:robot_layers}}
	\hfil{}
	\subfloat[A GRITSBot attached to the charging strips of the Robotarium.]{\includegraphics[width=0.25\textwidth]{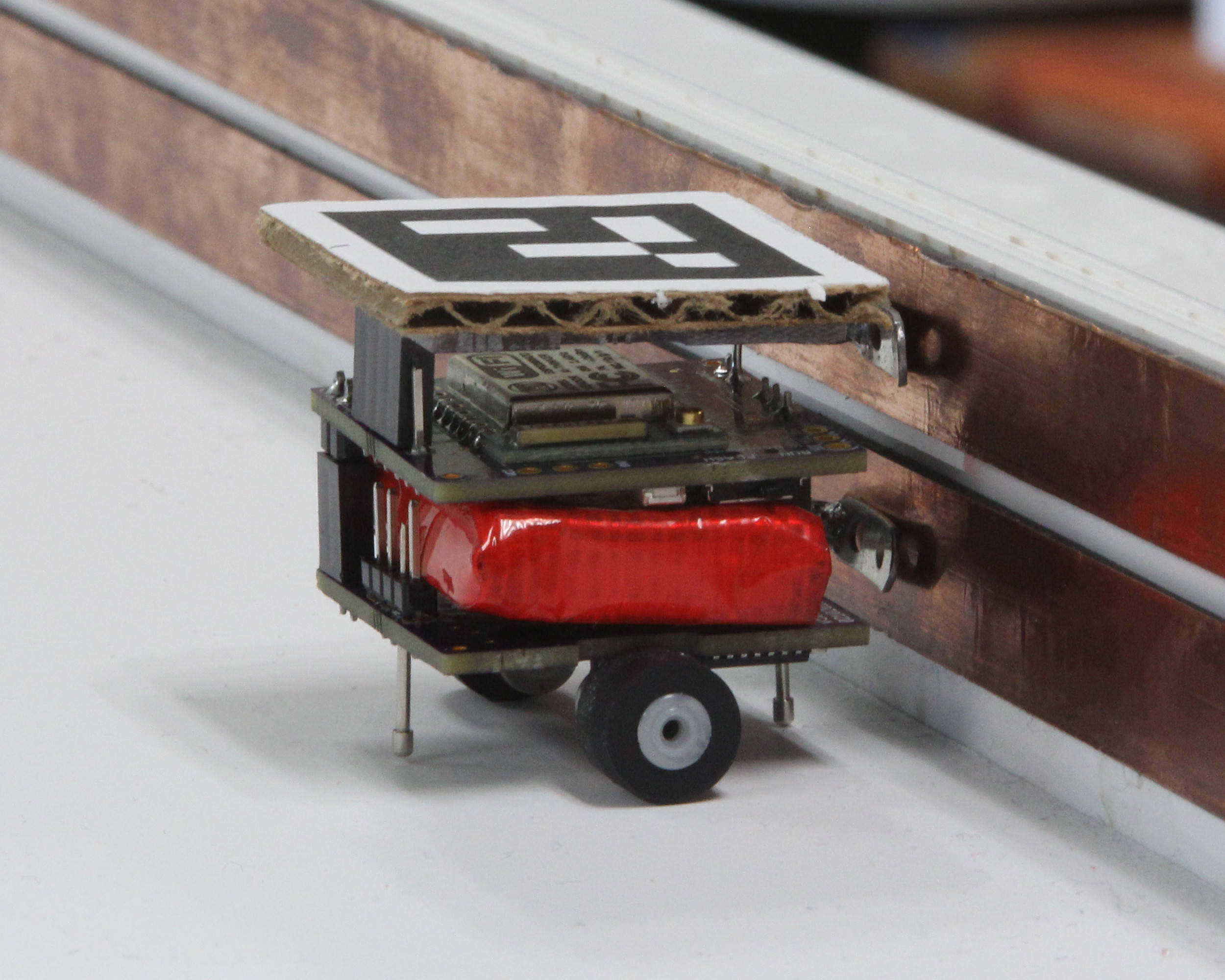} \label{fig:robot_charging}}
	\caption{Views of the GRITSBot, the mobile robot used in the Robotarium.\label{fig:robots}}
\end{figure}
    
\item{\textit{Tracking:}}
The global position of all robots is retrieved through an overhead tracking system and is required to close the position control feedback loop. The Robotarium uses a single webcam in conjunction with ArUco tags for tracking (see \cite{Garrido-Jurado2014}).
\footnote{ArUco is a minimal library for Augmented Reality applications based on OpenCV and can be found at \url{http://www.uco.es/investiga/grupos/ava/node/26}.} Note that most decentralized algorithms do not rely on global position updates but rather sensor data. However, system maintenance such as recharging robots automatically or setting up an experiment (i.e. moving robots to user-specified positions) relies on globally accurate position data. As such, overhead tracking is key to the robust operation and maintenance of the Robotarium.
      
\item{\textit{Charging:}}
Arguably the most crucial component of a self-sustaining and maintenance-free testbed is an automatic recharging mechanism for the robots. The GRITSBot has been designed for autonomous recharging through two extending prongs at the back of the robot that can connect to magnetic charging strips built into the testbed walls. This setup together with global position control enabled by overhead tracking allows the GRITSBots to autonomously recharge its battery (see Fig. \ref{fig:robot_charging}).
In the larger context of remotely accessible testbeds that contain mobile robots, the charging behavior is the key aspect of the GRITSBot that will enable automated use of the robots and management of the Robotarium hardware without operator intervention. The Robotarium back end assigns a number of available robots to users and ensures that these robots have been charged before the start of an experiment. After the conclusion of a user's experiment, robots are automatically returned to the charging station and prepared for the next user. Automating the recharging of robots is essential to making the continuous operation of the Robotarium economically feasible, especially as the number of robots is scaled up.
\end{itemize}

\subsubsection{Software}
\label{subsubsec:software}
The software components managing the operation of the Robotarium can be broadly grouped into three categories: coordinating server applications, components enabling the interaction with the testbed, and simulation-based components. This categorization can also be seen in the architecture diagram in Fig. \ref{fig:system_architecture}.
  
\begin{itemize}
\item{\textit{Simulation:}}
The simulation capabilities of the Robotarium are leveraged in three distinct ways: prototyping of user code, verification of user-provided code, and adding virtual robots. The simulator enable users to prototype and test their algorithms on their own machines before submitting them for execution on the Robotarium.\footnote{The simulator is currently implemented in Matlab and available at \url{www.robotarium.org}} Once submitted, but before being executed on the testbed, the same simulation infrastructure verifies collision-freedom (see Section \ref{subsec:sim_based_verification}). Additionally, the simulator also allows adding virtual robots to the testbed arena that are capable of interacting with the physical robots through the server back end.
    
\item{\textit{Interaction:}}
These components govern how users can interact with the Robotarium. Two principal methods of interaction are enabled by the components shown in Fig. \ref{fig:system_architecture}: local access through provided APIs as well as remote interaction through web-based code upload.\footnote{Note that no real-time teleoperation of robots is enabled for security reasons. Submitted user-code is executed locally on the Robotarium server and as such latency is negligible.} On the one hand, local API-based access requires users to connect to the same WiFi network as the server and the robots, which minimizes latency and enables closing the velocity or position control feedback loop through a user machine.
Remote access, on the other hand, requires users to implement their algorithms and test them in simulation before submitting them to the Robotarium via its web interface. Submitted code undergoes simulation-based verification that checks for error- and collision-free execution before code is executed on physical robots. Simulation-based verification is used as a stepping stone towards formal verification methods that are still in an active stage of development \cite{Roozbehani2005,Roozbehani2013}.\footnote{Remote-access to the Robotarium requires manual screening of applicants. Interested users can apply for time on the Robotarium via the website \url{www.robotarium.org}.}
  
\item{\textit{Coordination:}}
The server application is the central coordinating instance responsible for executing user code, routing commands and data to and from robots, transmitting global position data to the robots if needed, and managing simulated virtual robots. In addition the server logs all generated data and makes recorded videos and robot data available to users.
\end{itemize}

\subsection{Access through Abstractions}
\label{subsec:access_through_abstractions}
The Robotarium aims to support and promote research at many levels.  A mechanical engineer may access individual wheel velocities of a given robot.  At another, a middle school or high school student may prefer to issue high-level position commands (e.g., to navigate a virtual obstacle course).  Finally, a swarm roboticist may wish to control the Roboatrium's agents through dynamical or topological abstractions.  To enable these approaches and more, the Robotarium exposes access at various levels of abstractions, from wheel velocities to composable, swarm behaviors.  In particular, the Robotarium supports the following levels of access and abstraction: 

\begin{enumerate}
	\item \textit{Velocity}:  At the most basic level, users may implement individual wheel velocities for the differential drive robot.  At a slightly higher level, they can utilize a unicycle model, issuing linear and rotation velocity commands.
    \item \textit{Position}:
    	\begin{itemize}
        	\item Go-to-goal:  At this level, users may issue high-level, goal position commands or waypoints to the robots.
            \item Difference equations:  Researchers utilizing discrete-time dynamics may prefer to send "next position" commands to the agents, rather than velocity commands.
        \end{itemize}
    \item \textit{Modeling}: Swarm robotics algorithms typically abstract robot dynamics as single-integrators.  However, most real, robotic systems (e.g., the GRITSbot), due to their non-holonomic nature, cannot accomodate this abstraction directly.  The Robotarium offers diffeomorphisms, as in \cite{Olfati-Saber2002}, to translate between these dynamical models.
    \item \textit{Group control level}
    \begin{itemize}
    	\item Topological abstractions: At this level, users don't control individual robots anymore but command them to assemble certain topologies, formations, or shapes. At this level users have some control over the number of robots in their formations but not necessarily over individual agents. An example of interaction at this level is shown in Section \ref{subsec:example_coverage_control}.
        \item Behavioral abstractions: This level of abstraction is appropriate for users who care about the behavior that individual robots execute but only to a certain degree about how many robots there are in total. Robots may be interchangeable or execute distinct behaviors. The robots' dynamics and low-level control are entirely abstracted away for the user. An example of interaction at this level is shown in Section \ref{subsec:example_rendezvous})
    \end{itemize}
    \item \textit{Swarm abstractions}: At this highest level of abstraction, users control entire collectives of robots without any regard to individuals in the swarm. Neither the robots' dynamics nor the individual robots are of importance to the user. At this point the underlying hardware is completely abstracted away. An example of this interaction level is shown in Section \ref{subsec:example_coverage_control}.
\end{enumerate}

\section{Safety}
\label{sec:safety}
Allowing remote users to take control of the Robotarium's physical equipment imposes inherent risks to the integrity and safety of the hardware. To ensure safety and avoid damaging hardware, a combination of offline simulation-based verification and online collision avoidance using barrier certificates is employed.
By default, the execution of all user-supplied control code will be safe-guarded using barrier functions (see Section \ref{subsec:safety_barrier_certificates}). However, users can bypass this online safety mechanism to execute the raw user algorithm. Disabling barrier functions requires the user code to achieve a high safety score during the offline simulation-based verification step (as will be shown in Section \ref{subsec:sim_based_verification}). This safety score is a measure of how likely an algorithm will cause collisions during runtime. In that sense, a high safety score indicates low collision probabilities.


To introduce the two safety components in this section, we will rely on a commonly used dynamics model called the single integrator dynamics.
\begin{equation}\label{eqn:single_integrator}
	\dot{x}_i = u_i.
\end{equation}
Here, $x_i\in\mathbb{R}^2$ is the position of robot $i$ in the plane and $u_i\in\mathbb{R}^2$ is its velocity input.\footnote{Note that even though we consider the 2D case here, the methods shown in this section also extend to 3D applications.} The reason for this choice is twofold. First, the Robotarium supports the single integrator abstraction (according to Section \ref{subsec:access_through_abstractions} - modeling level), and, second, control algorithms for multi-robot systems are often developed using the single integrator model, e.g., \cite{Jadbabaie2003, Ji2007, Cortes2002}. Although the GRITSBot is a unicycle-type robot, the default velocity controller - a diffeomorphism controller - maps these unicycle dynamics to single integrator dynamics for general usage (see \cite{Olfati-Saber2002}). Despite the focus in this section on single integrator dynamics, it is important to note that numerous other abstractions are supported according to \ref{subsec:access_through_abstractions}.

\subsection{Simulation-based Code Verification}
\label{subsec:sim_based_verification}
The simulation environment provided for the user to prototype algorithms also provides a platform for characterizing the safety of the controller.  A two-phase collision detection algorithm, similar to those described in \cite{mirtich1997,Klancar2007}, is used to determine the time and location of contact occurrences between objects. Once contact is detected, the simulator enforces a non-penetration constraint \cite{baraff1989} and records the command and state data of the robot(s) involved. The velocities and relative geometry at instances of contact are used to generate a safety score $S$ for the controller, which is unity for the collision-free case and zero for the theoretical worst case of collision of all robots at every iteration, each with their maximum velocity commanded directly into the contact surface.


  
To further aid in the prototyping process, detailed feedback on the controller's performance is returned to the user (in addition to the safety score itself). Parameters such as the mean collision velocity, or mean contact duration may be useful in diagnosing the cause of low safety scores. 
The main purpose of the safety function is to determine whether a given controller is likely to be executable in a safe fashion on the Robotarium. If $S$ lies above a certain threshold value, the controller is deemed to be safely executable without any further safety measures. If the score $S$, however, is too low, users will be given the options to refine their code or to accept safety barrier function wrapping, which will result in provably collision-free code execution. In the next section, we will introduce safety barrier certificates and elaborate on how they can be deployed in conjunction with the user code in a minimally invasive fashion.


\begin{figure*}[t]
	\centering
	\subfloat[Time 2.0s]{\includegraphics[width=0.32\textwidth]{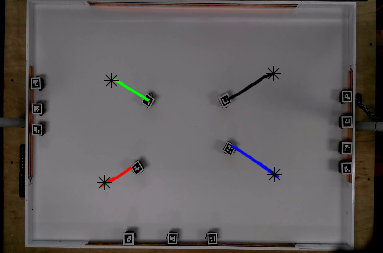} \label{fig:exp1}}
	\subfloat[Time 4.0s]{\includegraphics[width=0.32\textwidth]{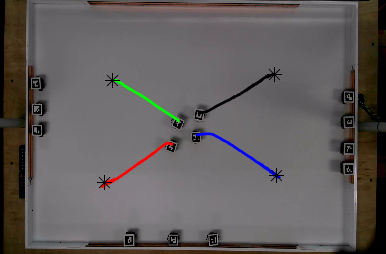} \label{fig:exp2}}
	\subfloat[Time 6.0s]{\includegraphics[width=0.32\textwidth]{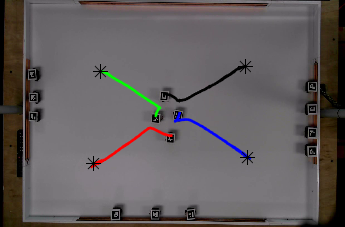} \label{fig:exp3}} \\
	\subfloat[Time 7.5s]{\includegraphics[width=0.32\textwidth]{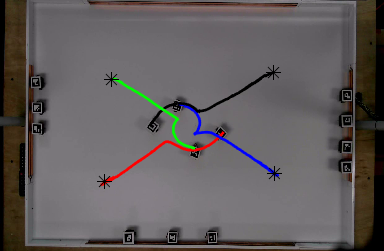} \label{fig:exp4}}
	\subfloat[Time 10.3s]{\includegraphics[width=0.32\textwidth]{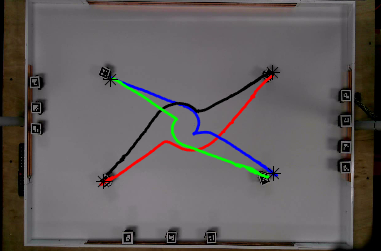} \label{fig:exp5}}
	\caption{Four GRITSBots swap positions with active safety barrier certificates on the Robotarium. The curves are the trajectories of the robots overlaid on the image. The star markers represent the initial positions of the robots. \label{fig:expbarrier}}
\end{figure*}

\subsection{Safety Barrier Certificates}
\label{subsec:safety-barrier-certificates}
\label{subsec:safety_barrier_certificates}
The Robotarium uses \textit{Safety Barrier Certificates} to ensure provably collision-free behavior of all robots. This multi-robot collision avoidance strategy is based on the following three principles.
\begin{itemize}
  \item All robots are provably safe in the sense that collision are avoided.
  \item Users' commands are only modified when collisions are truly imminent.
  \item Collision avoidance is executed in real-time (up to 20 Hz update rate).
\end{itemize} 
These safety barrier certificates are enforced through the use of
control barrier functions, which are Lyapunov-like functions that can provably guarantee forward set invariance, i.e. if the system starts in the safe set, it stays in the safe set for all time. A specific class of maximally permissive control barrier functions were introduced in \cite{xu2015, ames2014}, whose construction provides the basis for the minimally invasive safety guarantees afforded by the Robotarium.

Consider a team of $N$ mobile robots with the index set $\mathcal{M}=\{1,2,...,N\}$. Each robot $i$ uses single integrator dynamics according to \eqref{eqn:single_integrator}.  
Additionally, robot $i$'s velocity $u_i$ is bounded by $\|u_i\|\leq \alpha, \forall i\in\mathcal{M}$. Let $x=[x_1^T, x_2^T, ..., x_N^T]^T$ and $u=[u_1^T, u_2^T, ..., u_N^T]^T$ denote the aggregate state and velocity input of the entire team of robots. 

To avoid inter-robot collisions, any two robots $i$ and $j$ need to maintain a minimum safety distance $D_s$ between each other. This requirement is encoded into a pairwise safe set $\mathcal{C}_{ij}$, which is a super level set of a smooth function $h_{ij}(x)$, 
\begin{equation}\label{eqn:setcij}
\mathcal{C}_{ij} = \{x_i\in\mathbb{R}^2 ~|~h_{ij}(x)=\|x_i-x_j\|^2-D_s^2\geq 0\}, ~\forall~ i\neq j.
\end{equation}

The function $h_{ij}(x)$ is called a control barrier function, if the admissible control space
\begin{equation}\label{eqn:setkij}
K_{ij}(x) = \left\{u\in\mathbb{R}^{2N} \;\middle|\; \frac{\partial h_{ij}(x)}{\partial x}u\geq -\gamma h_{ij}(x)\right\},
\end{equation}
is non-empty for all $x_i\in\mathcal{C}_{ij}$. Note that $u=0$ is always in $K_{ij}(x)$. It has been shown in \cite{xu2015} that if the control input $u$ stays in $K_{ij}(x)$ for all time, then the safe set $\mathcal{C}_{ij}$ is forward invariant. In addition, the forward invariance property of $\mathcal{C}_{ij}$ is robust with respect to different perturbations on the system. 

Combining \eqref{eqn:setcij} and \eqref{eqn:setkij} as well as the single integrator dynamics in \eqref{eqn:single_integrator}, the velocity input $u$ needs to satisfy
\begin{equation*}
-2(x_i-x_j)u_i+2(x_i-x_j)u_j\leq \gamma h_{ij}(x),~ \forall~ i\neq j.
\end{equation*}
This inequality can be treated as a linear constraint on $u$ when the state $x$ is given, i.e., 
\begin{equation*}
A_{ij}u\leq b_{ij}, ~\forall~ i\neq j,
\end{equation*}
where 
\begin{eqnarray}
	A_{ij} &=&[0, \ldots, \underbrace{-2(x_i-x_j)^T}_{\text{robot} ~i}, \ldots, \underbrace{2(x_i-x_j)^T}_{\text{robot} ~j}, \ldots, 0 ] \nonumber\\ 
    b_{ij} &=& \gamma h_{ij}(x) \nonumber
\end{eqnarray}
In addition to inter-robot collisions, collisions with the workspace boundaries also need to be avoided. Assume the workspace of the Robotarium is bounded by a rectangle $[B_l, B_r, B_b, B_t]$ (left, right, bottom, and top bounds), i.e., $x_i$ needs to stay in the set
\begin{eqnarray*}
\bar{\mathcal{C}}_i = \{  x_i\in\mathbb{R}^2 ~|~ \bar{h}_{i1}(x) =(B_r-x_i[1])(x_i[1]-B_l)\geq 0, \\ \bar{h}_{i2}(x) = (B_t-x_i[2])(x_i[2]-B_b)\geq 0, \}
\end{eqnarray*}
where $x_i=[x_i[1],x_i[2]]^T$. As before, safety barrier constraints are synthesized to ensure the forward invariance of $\bar{\mathcal{C}}_i$,
\begin{equation*}
\bar{A}_{i}u_i\leq \bar{b}_{i}, ~\forall~i\in\mathcal{M},
\end{equation*}
where 
\begin{eqnarray}
	\bar{A}_{i}&=&\begin{bmatrix} 2x_i[1]-B_r-B_l&0\\ 0&2x_i[2]-B_t-B_b \end{bmatrix} \nonumber\\
    \bar{b}_{i}&=&\begin{bmatrix} \gamma\bar{h}_{i1}(x)\\ \gamma\bar{h}_{i2}(x) \end{bmatrix} \nonumber
\end{eqnarray}
To ensure the safety of the entire robot team, \emph{all} pairwise collisions and collisions with the workspace boundaries need to be excluded, which can be encoded as,
\begin{equation*}
\mathcal{C} = \underset{{i \in \mathcal{M}}}{\prod} \Big\{{\bigcap_{\substack{j \in \mathcal{M}\\ j\neq i}}\mathcal{C}_{ij} \bigcap\bar{\mathcal{C}}_{i} }\Big\}, 
\end{equation*}
where $\underset{{i \in \mathcal{M}}}{\prod}$ is the Cartesian product across all robots' states. The forward invariance of the safe set $\mathcal{C}$ is guaranteed by the safety barrier certificates, which are defined as
\begin{equation}\label{eqn:certificates}
K(x) = \left\{u\in\mathbb{R}^{2N}\;\middle|\;A_{ij}u\leq b_{ij}, \bar{A}_{i}u_i\leq \bar{b}_{i}, ~\forall~ i\neq j \right\}.
\end{equation}
These safety barrier certificates define a convex polytope $K(x)$ in which users' control commands shall always stay. By constraining users' control commands to within $K(x)$, the Robotarium is guaranteed to operate in a provably collision-free manner.
Note that these safety barrier certificates are deployed in conjunction with the user-specified controller. The \textit{minimally invasive} nature of  barrier certificate-enabled collision avoidance stems from the fact that the deviation between the user-specified control signal and the actually executed signal is minimized, subject to safety constraints. This minimally invasive safety guarantee is realized through a Quadratic Program (QP)-based controller
\begin{equation*}
\label{eqn:QPcontroller}
 \begin{aligned}
u^* =  & \:\: \underset{u\in\mathbb{R}^{2n}}{\text{argmin}}
 & & J(u) = \sum_{i=1}^{N} \|{u}_{i} - \hat{u}_{i} \|^2 &\\
 & \qquad \text{s.t.}
 & & A_{ij}u \leq b_{ij}, \qquad  &\forall~ i\neq j,  &  \\
 &
 & & \bar{A}_{i}u_i\leq \bar{b}_{i},\: \qquad  &\forall~ i\in \mathcal{M}, & \\
 &
 & &    \| u_i\|_\infty  \leq \alpha, \qquad &\forall~ i \in\mathcal{M},
 \end{aligned}
\end{equation*}
where $\hat{u}$ is the user's control command and $u^*$ is the actually executed control command. Note that in the absence of impending collisions (i.e. when the safety barrier certificates in \eqref{eqn:certificates} are satisfied), the user's code is executed faithfully. When violations occur, a closest possible (in a least-squares sense) safe control command is computed and executed instead. The online QP-based controller can be executed in real-time on the Robotarium (with an update rate of approximately 20 Hz).
An experiment showing four GRITSBots swapping positions with active safety barrier certificates is shown in Fig. \ref{fig:expbarrier}. A link to the video of this experiment can be found in Table \ref{table:multimedia_extensions}.

\section{Examples}
\label{sec:experiments}
In this section we present a number of canonical examples of networked control algorithms that have been instantiated on the Robotarium. Specifically, we elaborate on the rendezvous problem, formation control, and the coverage control problem. Besides the mathematical introduction of each algorithm, we show code samples and detailed results of executing these examples on the Robotarium. Additionally, links to multimedia extensions showing these algorithms in action can be found in Table \ref{table:multimedia_extensions}. 
The examples in this section serves two primary purposes. On the one hand they showcases the breadth of algorithms that can be executed on the Robotarium. On the other hand, they serve as a brief tutorial demonstrating how easily swarm-robotic algorithms can be instantiated using the Robotarium's infrastructure. In particular, in this section, we will make use of a Matlab-based simulator and a Matlab API to communicate with the Robotarium backend (these components have been outlined in Section \ref{subsubsec:software} and are shown in Fig. \ref{fig:system_architecture}.\footnote{The simulator required to run these code samples can be downloaded at \url{https://github.com/robotarium/robotarium-matlab-simulator}.}
%

\subsection{Rendezvous}
\label{subsec:example_rendezvous}

The rendezvous problem (or consensus) is a canonical example in networked control theory with applications ranging from sensor fusion to swarm robotics.  Within the area of swarm robotics, the premise is simple: using only local information, a group of mobile agents must meet at the same location.

This algorithm employs a team of $N$ single-integrator agents, as defined in \eqref{eqn:single_integrator}.  Let the agents' static communication network be described by an undirected graph $G = (V, E)$, where $(i, j), (j, i) \in E$ implies that agents $i$ and $j$ share information.  A common solution to the rendezvous problem is to let $u_{i}$ be defined using the local interaction rule (e.g., as in \cite{Jadbabaie2003})
\begin{equation}
    \dot{x}_{i} = \sum_{j \in N_{i}} (x_{j} - x_{i})
    \label{eqn:node-level-dynamics}
\end{equation}
where $N_{i}$ represents the neighbors of agent $i$ induced by $G$. The translation of this formal mathematical description into deployable code is shown in Fig.~\ref{fig:consensus-translation}. In particular, Fig.~\ref{fig:consensus-translation} shows an implementation of \eqref{eqn:node-level-dynamics} using the Robotarium's MATLAB API.
\begin{figure}[tbp]
    \centering
    \lstset{language=Matlab,%
		basicstyle=\small,%
        breaklines=true,%
        morekeywords={matlab2tikz},%
        keywordstyle=\color{blue},%
        morekeywords=[2]{1}, keywordstyle=[2]{\color{black}},%
        identifierstyle=\color{black},%
        stringstyle=\color{mylilas},%
        commentstyle=\color{mygreen},%
        frame=single,%
        showstringspaces=false,%
        emph=[1]{for,end,break},emphstyle=[1]\color{blue},%
    }
\begin{lstlisting}[language=Matlab]
r = RobotariumMatlabAPI();
G = communicationTopology;

for i = 1:N
    dx(:, i) = [0 ; 0];
    x = r.getPoses();
    
    for j =  r.getTopNeighbors(i, G);
        
        dx(:, i) = dx(:, i) + ...
                   (x(1:2, j) - x(1:2, i));
    
    end
    
    r.setVelocities(dx);
end
\end{lstlisting}
    \caption{Code example of the rendezvous controller defined in \eqref{eqn:node-level-dynamics} using the Robotarium's MATLAB API.}
    \label{fig:consensus-translation}
\end{figure}

Note that this algorithm utilizes a modeling abstraction, as detailed in Section~\ref{subsec:access_through_abstractions}; so, we must also map the generated single-integrator control inputs to the unicycle dynamics of the GRITSbot, which is described in the full example code and documentation referenced in Table \ref{table:code_samples}. 

\begin{figure}[b]
    \centering
    \includegraphics[width=0.5\textwidth]{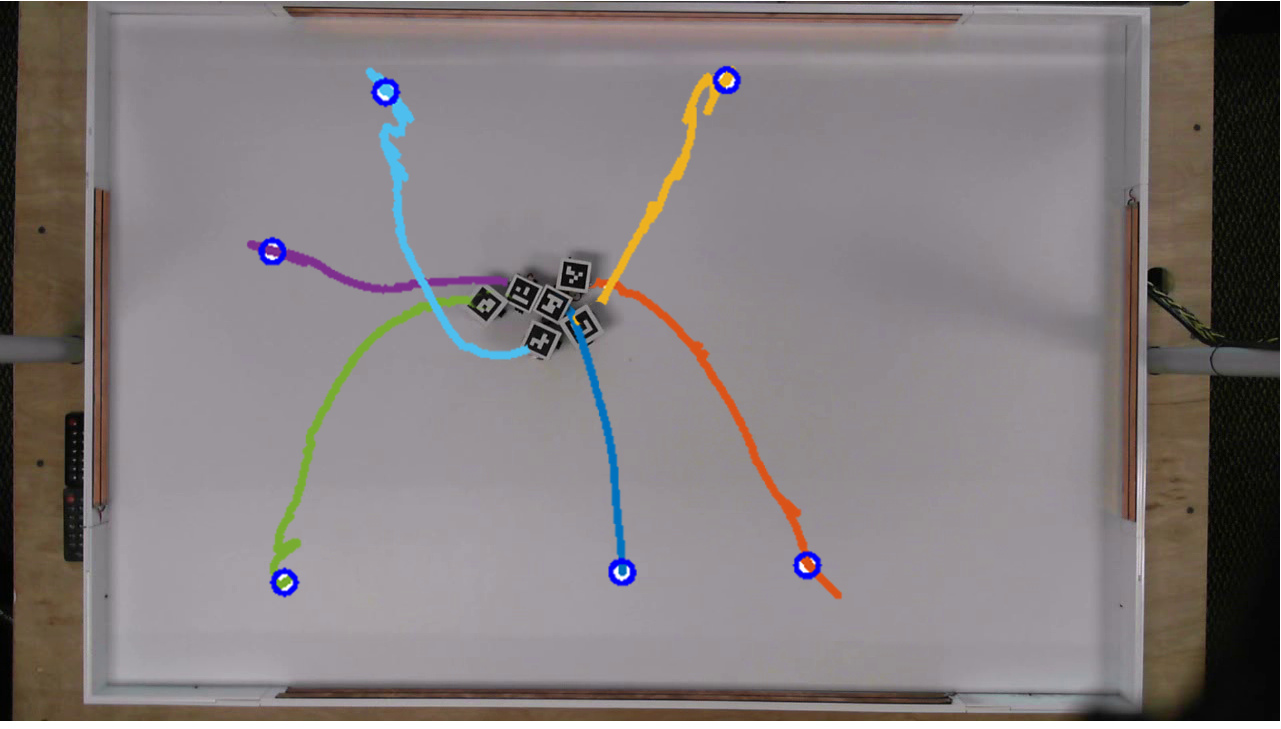}
    \caption{GRITSbot trajectories during the deployment of the consensus algorithm onto the Robotarium.}
    \label{fig:consensus-data}
\end{figure}

For our particular experiment, we selected $N = 6$ and $G = C_{6}$ (i.e., an undirected, connected, cycle graph containing 6 agents).  Deploying the previously shown code onto the Robotarium yielded the results shown in Fig.~\ref{fig:consensus-data}.  Table~\ref{table:multimedia_extensions} contains a video of the robots' execution of this algorithm.

\subsection{Formation Control}
\label{subsec:example_formation_control}
The goal of formation control is to drive the agents to a configuration that satisfies a set of given constraints (i.e., desired inter-agent distances).  A variety of application domains utilize formation control, from sensor networks to convoy protection, and we here illustrated the operation of the Robotarium on a particular choice of algorithm given in \cite{Ji2007}.

As in the previous example, this algorithm employs a team of $N$ single-integrator agents, as defined in \eqref{eqn:single_integrator}. Assume that the underlying, static information exchange topology is rigid, and define a so-called edge-tension function (\cite{Ji2007}) as
\[
w(x) = \dfrac{1}{2}\sum_{i = 1}^{N}\sum_{j \in N_{i}} w_{ij}(x).
\]
Then,
\[
u_{i} = -\dfrac{\partial w(x)}{\partial x_{i}}
\]
can be used to drive the agents to the desired configuration, with the proper choice of $w_{ij}(x)$.  
Let for example $w_{ij}$ be defined as 
\[
w_{ij}(x) = \dfrac{\alpha}{4}(\|x_{i} - x_{j}\|^{2} - d_{ij}^{2})^{2}
\]
where $d_{ij} \in \mathbb{R}^{+}$ is the desired distance between two agents, and $\alpha \in \mathbb{R}^{+}$ is a gain.  The gradient of this function is
\[
\dfrac{\partial w(x)}{\partial x_{i}} = -\sum_{j \in N_{i}} \alpha(\|x_{i} - x_{j}\|^{2} - d_{ij}^{2})(x_{j} - x_{i}),
\]
which results in
\begin{equation}
    \dot{x}_{i} = \sum_{j \in N_{i}} \alpha(\|x_{i} - x_{j}\|^{2} - d_{ij}^{2})(x_{j} - x_{i}).
    \label{eqn:formation-control-law}
\end{equation}
Fig.~\ref{fig:formationControl-translation} shows the implementation of \eqref{eqn:formation-control-law} using the Robotarium's MATLAB API. 
\begin{figure}[t]
    \centering
    \lstset{language=Matlab,%
        basicstyle=\small,%
        breaklines=true,%
        morekeywords={matlab2tikz},%
        keywordstyle=\color{blue},%
        morekeywords=[2]{1}, keywordstyle=[2]{\color{black}},%
        identifierstyle=\color{black},%
        stringstyle=\color{mylilas},%
        commentstyle=\color{mygreen},%
        frame=single,%
        showstringspaces=false,%
        emph=[1]{for,end,break},emphstyle=[1]\color{blue},%
    }
\begin{lstlisting}[language=Matlab]
r = RobotariumMatlabAPI();
G = rigidTopology;

for i = 1:N
    dx(:, i) = [0 ; 0];
    x = r.getPoses();
    
    for j = r.getTopNeighbors(i, G) 
        dx(:, i) = dx(:, i) + alpha * ...
          (norm(x(1:2, i) - x(1:2, j))^2 - ...
          weights(i, j)^2) * ...
          (x(1:2, j) - x(1:2, i));
    end 
    
    r.setVelocities(dx);
end
\end{lstlisting}
    \caption{Code example of the formation controller defined in \eqref{eqn:formation-control-law} using the Robotarium's MATLAB API.}
    \label{fig:formationControl-translation}
\end{figure}
Similar to the previous example, this algorithm utilizes a modeling abstraction detailed in Section~\ref{subsec:access_through_abstractions}. As such, the generated single-integrator control inputs need to be translated to the unicycle dynamics of the GRITSbot, which is described in the full example code and documentation referenced in Table \ref{table:code_samples}.   

\begin{figure}[tbp]
    \centering
    \includegraphics[width=0.49\textwidth]{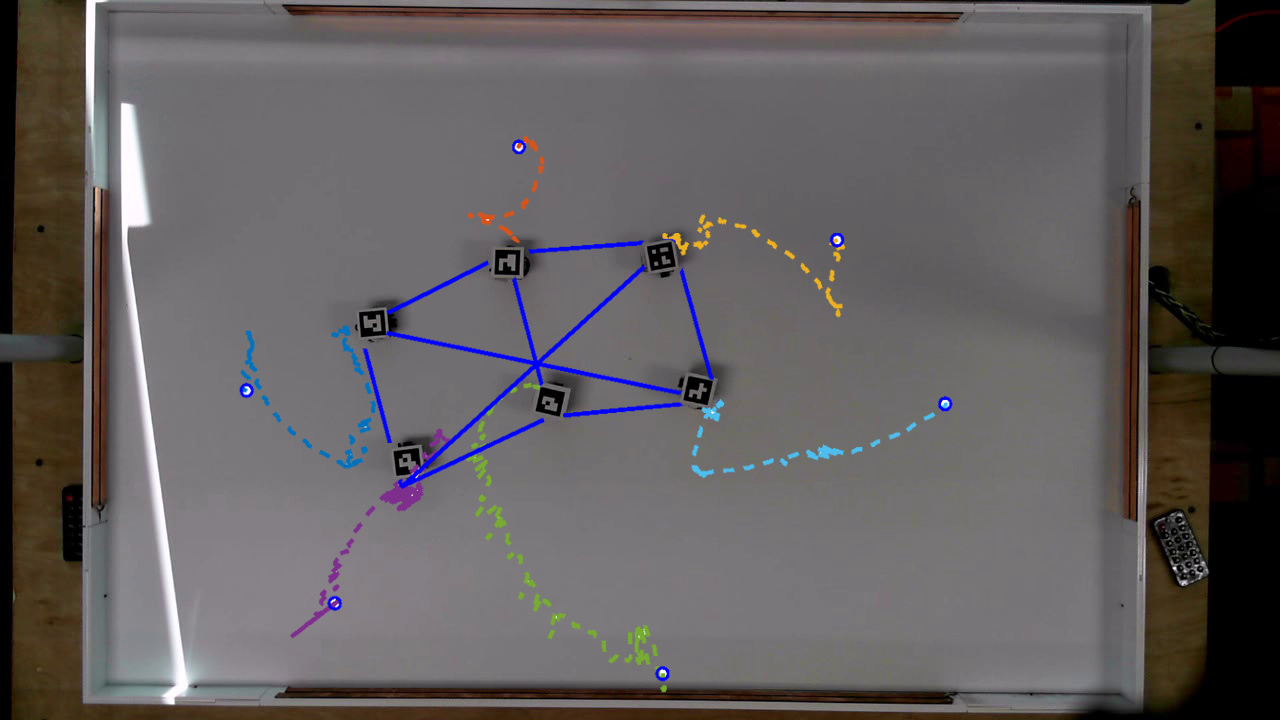}
    \caption{GRITSbot trajectories (dashed lines) and communication topology (solid lines) during the deployment of the decentralized formation control algorithm onto the Robotarium.}
    \label{fig:fc-data}
\end{figure}

We deployed this algorithm onto the Robotarium with $N = 6$ agents and a rigid formation specification as shown in Fig. \ref{fig:fc-data}.  Fig.~\ref{fig:fc-data} also visualizes the communication topology and the trajectories of the robots during the experiment. The seemingly erratic motion in some agents' trajectories stemmed from the rigidity of the formation, which restricted the agents' movements.  Table~\ref{table:multimedia_extensions} contains a video reference for this algorithm.
\subsection{Coverage Control}
\label{subsec:example_coverage_control}
The coverage control problem is another canonical example of multi-agent control that has been implemented on the Robotarium. 
Variations of the coverage problem have been extensively studied (e.g., see \cite{Du1999,Aurenhammer1991,Lloyd1982} and the references therein). Applications include data compression in image processing, quantization and clustering, optimal placement of resources, animal behavior description, and mobile sensor networks, to name a few.


Formally, we will consider a team of $N$ agents, whose positions within the domain $D$ are given by $x_i\in D$, $i \in\mathcal{M}$, and have single integrator dynamics as in \eqref{eqn:single_integrator}. Our criterion for coverage is captured by the locational cost \cite{Cortes2002}
\begin{align*}
\mathcal{H}(x,t) = \sum_{j\in\mathcal{M}}\int_{V_i(x)}\left\|q-x_i\right\|^2\phi(q,t) \,\mathrm{d}q
\end{align*}
where the domain $D$ has been partitioned into so-called regions of dominance. In particular, a Voronoi tessellation of the space is used, i.e.,
\begin{align*}
V_i(x) = \left\{q\in D\;\middle|\;\left\|q-x_i\right\|\leq\left\|q-x_j\right\|, ~j\in\mathcal{M} \right\}.
\end{align*}
We will think of the time-varying density function $\phi: D\times[0,\infty)\to\left(0,\infty\right)$ as a human-generated input to the robot team, which captures the relative importance of points within the domain at any given time.

The following distributed control law, originally presented in \cite{Lee2015}, will achieve coverage
of the provided time-varying density function
\begin{align}\label{eqn:TVD-D1}
u_i = \kappa\left(c_i - x_i\right)+\frac{\partial c_i}{\partial t} + \sum_{j\in\overline{\vphantom{\vert}\mathcal{N}}_i} \frac{\partial c_i}{\partial x_j}\left(\kappa\left(c_j - x_j\right)+\frac{\partial c_j}{\partial t}\right)
\end{align}
for some $\kappa>0$, where $\overline{\vphantom{\vert}\mathcal{N}}_i$ is agent $i$'s closed neighborhood set in the Delaunay graph, the dual to the Voronoi tessellation generated by the agents' position in the domain, and $c_i$ is the center of mass of agent $i$'s Voronoi cell (for the definition of $c_i$ and the derivation of these partial derivatives, see \cite{Diaz-Mercado2015}).

Note that \eqref{eqn:TVD-D1} assumes the agents are modeled using single integrator dynamics and that agents are able to accelerate instantaneously to the required velocities. However, such conditions are rarely encountered in practice. In order to close the theory-practice gap, control law \eqref{eqn:TVD-D1} was implemented on the Robotarium. The Robotarium's flexibility allowed for a number of different input modalities to be used to generate a human-provided density functions: a touchscreen interface (e.g., a tablet device), hand-tracking and gesture recognition (e.g., via a Leap Motion controller), and a brain-computer interface (e.g., EEG). With all these input modalities, the user-provided reference (representative of the areas of interest within the domain) were transmitted using the TUIO protocol\footnote{TUIO is an open framework that defines a common protocol and API for tangible multitouch surfaces. See \url{tuio.org}.} to the Robotarium server. 
Density functions were generated from these reference locations as discussed in \cite{Diaz-Mercado2015,Diaz-Mercado2016}. Fig. \ref{fig:CoverageMatlabCode} illustrates the basic structure of the Matlab code used to run coverage control on the Robotarium, where the functions not in the Robotarium API were provided by the user.
\begin{figure}[tbp]
    \centering
    \lstset{language=Matlab,%
        basicstyle=\small,
        breaklines=true,%
        morekeywords={matlab2tikz},
        keywordstyle=\color{blue},%
        morekeywords=[2]{1}, keywordstyle=[2]{\color{black}},
        identifierstyle=\color{black},%
        stringstyle=\color{mylilas},
        commentstyle=\color{mygreen},%
        frame=single,%
        showstringspaces=false,%
        emph=[1]{for,end,break},emphstyle=[1]\color{blue},%
    }
\begin{lstlisting}[language=Matlab]
r = RobotariumMatlabAPI();
N = r.N(); D = r.getDomain();
x = r.getPoses();[phi,dphidt]= getDensity();
k = 1; % Control tuning parameter.
VCells = voronoiTessellation(x,D);
dx = zeros(2,N);
for ii = 1:N
    dcidxi = zeros(2);
    [ci,dcidt] = ...
     areaIntegrals(x(:,ii),VCells,phi,dphidt);
    for jj = getDelaunayNeighbors(ii,VCells)
        [cj,dcjdt] = areaIntegrals(x(:,jj),...
                         VCells,phi,dphidt);
        [dcidxij,dcidxj] = lineIntegrals(...
                  x(:,ii),x(:,jj),VCells,phi);
        dcidxi = dcidxi + dcidxij;
        dx(:,ii) = dx(:,i) + ...
            dcidxj*(k*(cj-x(:,jj))+dcjdt);
    end
    dx(:, ii) = dx(:, ii) + ...
       (eye(2)+dcidxi)*(k*(ci-x(:,ii))+dcidt);
end
r.setVelocities(dx);
\end{lstlisting}
    \caption{Matlab code used for coverage control on the Robotarium as depicted in Fig. \ref{fig:static_coverage_control_title_page} and Fig. \ref{fig:LeapMotion}.}
    \label{fig:CoverageMatlabCode}
\end{figure}

\begin{figure}[tbp]
\centering
\includegraphics[width=\linewidth, clip=true, trim= 200 0 300 100]{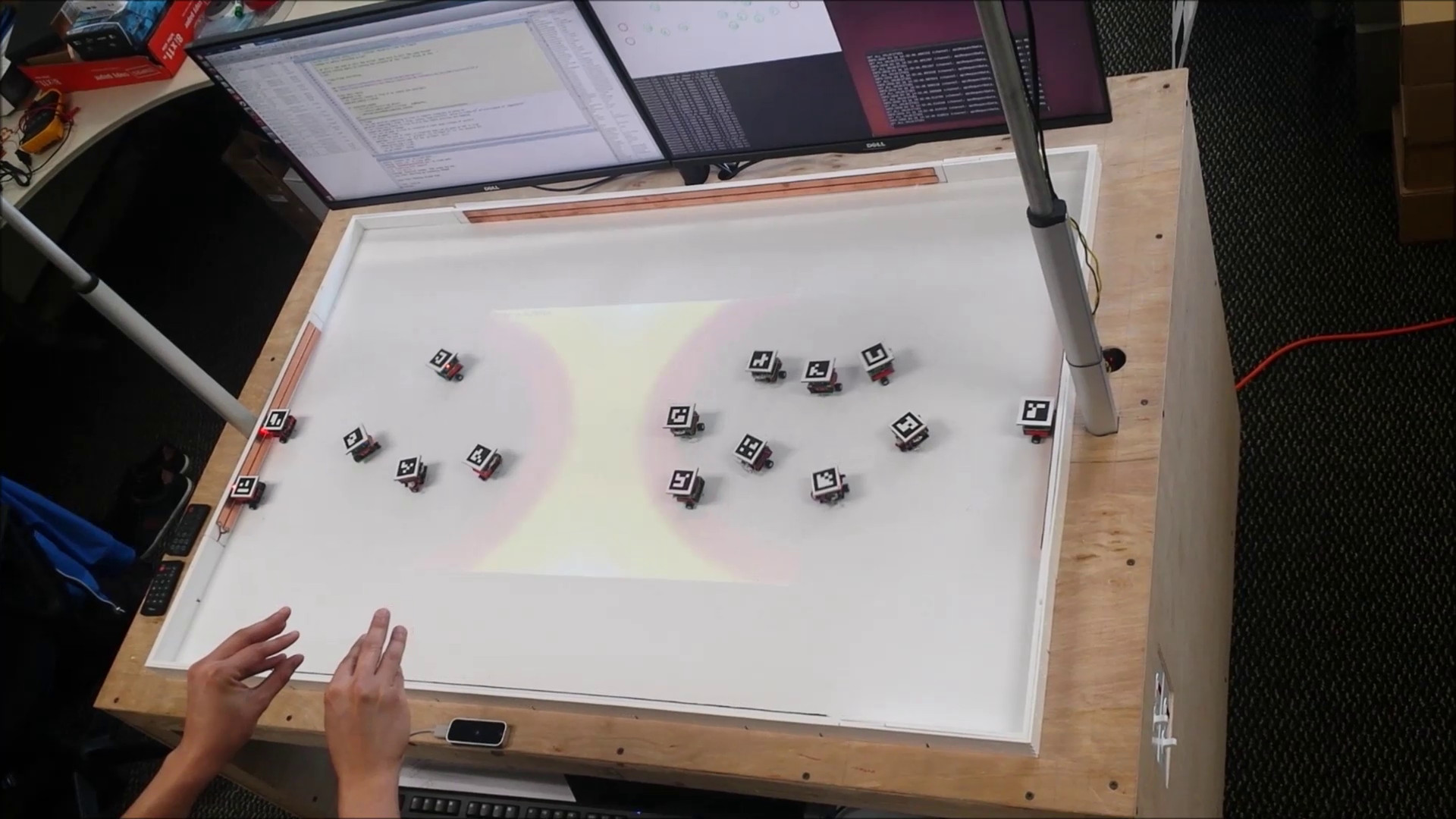}
\caption{A Leap Motion controller is used to generate densities for coverage control.
}\label{fig:LeapMotion}
\end{figure}

Fig. \ref{fig:LeapMotion} illustrates a coverage control experiment carried out on the Robotarium using 12 GRITSBots. For these experiments, a Leap Motion controller was used to detect and track the human-operator's hands. The finger positions 
were used as reference locations of the areas of importance over the domain to generate the time-varying densities to be covered by the robot team. To provide visual feedback for the human operator, an overhead projector was used to visualize and overlay the generated density over the robot domain in real-time. In Fig. \ref{fig:LeapMotion}, we see the user effectively clustering the 12 active robots into two different groups at the desired locations, while three inactive robots charge. An overhead projector is used to visualize and overlay the human-provided density function on the domain in real-time.

\section{Conclusion}
\label{sec:conclusion}
In this paper, we have detailed the development of a remotely accessible, multi-robot research facility -- the \textit{Robotarium}. Beyond introducing the hardware and software components required to enable remote accessibility of swarms of robots, the Robotarium addressed the two key concerns of flexibility and safety. Unlike other remotely accessible testbeds, the Robotarium makes use of formal methods to ensure the safety of its physical assets and the avoidance of damage to the robots. These methods guarantee provable collision avoidance in a minimally invasive manner without overly constraining the type of control algorithms that can be executed on the Robotarium. To demonstrate the flexibility and versatility of this testbed, we have shown a number of examples that were deployed on the Robotarium with little implementation overhead.

While the current instantiation of the Robotarium is fully functional and remotely accessible at \url{www.robotarium.org}, future work includes increasing the number of available robots such that multiple experiments can be carried out in parallel, but also diversifying the types of available robots. Developments are underway to augment the current GRITSBot-based testbed with robots exhibiting different locomotion modalities and more complex dynamics such as quadcopter or bipedal humanoids.


\begin{table}[tbp]
  \centering
  \begin{tabular}{lll}
    \bfseries Extension & \bfseries Algorithm & \bfseries Link to Video \\
    \hline
    1 & Rendezvous 					& \url{https://youtu.be/mAmdrta8jio} \\
    \rowcolor{LightGray}
    2 & Formation Control 			& \url{https://youtu.be/nm4jUjTxZ_U} \\
    3 & Safety Barrier Certificates   & \url{https://youtu.be/E_Q8e1Adc48} \\
    &&\\
  \end{tabular}
  \caption{List of video references.}
  \label{table:multimedia_extensions}
\end{table}

\begin{table}[tbp]
  \centering
  \begin{tabular}{lll}
    \bfseries Extension & \bfseries Algorithm & \bfseries Link to Code Sample\\
    \hline
    1 & Rendezvous 			& \url{http://robotarium.github.io/examples/} \\
    \rowcolor{LightGray}
    2 & Formation Control 	& \url{http://robotarium.github.io/examples/} \\
    &&\\
  \end{tabular}
  \caption{List of code samples.}
  \label{table:code_samples}
\end{table}


\bibliographystyle{IEEEtran}
\bibliography{references}
%
%
%

\end{document}